\documentclass[sigconf]{acmart}

\usepackage{booktabs} 

\usepackage{bm}
\usepackage{amsmath}
\usepackage{dsfont}
\usepackage{amssymb}
\usepackage[gen]{eurosym}
\usepackage{makecell}
\usepackage{multirow}
\usepackage{algorithmic}
\usepackage{algorithm}
\usepackage{color}
\usepackage{balance}
\usepackage{enumitem}
\setitemize[1]{itemsep=0pt,partopsep=0pt,parsep=0pt,topsep=0pt}

\renewcommand{\baselinestretch}{0.985}

\begin{document}
\title{Multi-Label Image Classification via Knowledge Distillation from Weakly-Supervised Detection}
\author{Yongcheng Liu$^{1,2}$, \ \ Lu Sheng$^{3}$, \ \ Jing Shao$^{4,\mathbf{*}}$, \ \ Junjie Yan$^{4}$, \ \ Shiming Xiang$^{1,2}$, \ \ Chunhong Pan$^{1}$}
\affiliation{%
\institution{$^{1}$ National Laboratory of Pattern Recognition, Institute of Automation, \ Chinese Academy of Sciences}
}
\affiliation{\institution{$^{2}$ School of Artificial Intelligence, \ University of Chinese Academy of Sciences}}
\affiliation{\institution{$^{3}$ CUHK-SenseTime Joint Lab, \ The Chinese University of Hong Kong \ \ $^{4}$ SenseTime Research}}
\affiliation{\institution{\{yongcheng.liu, smxiang, chpan\}@nlpr.ia.ac.cn, \ \ lsheng@ee.cuhk.edu.hk, \ \ \{shaojing$^{\mathbf{*}}$, yanjunjie\}@sensetime.com}}
\affiliation{\textcolor[rgb]{1.00,0.00,1.00}{\texttt{https://yochengliu.github.io/MLIC-KD-WSD/}}}

\renewcommand{\shortauthors}{Liu et al.}

\begin{abstract}
Multi-label image classification is a fundamental but challenging task towards general visual understanding. Existing methods found the region-level cues (e.g., features from RoIs) can facilitate multi-label classification.
Nevertheless, such methods usually require laborious object-level annotations (i.e., object labels and bounding boxes) for effective learning of the object-level visual features.
In this paper, we propose a novel and efficient deep framework to boost multi-label classification by distilling knowledge from weakly-supervised detection task without bounding box annotations.
Specifically, given the image-level annotations, (1) we first develop a weakly-supervised detection (WSD) model, and then (2) construct an end-to-end multi-label image classification framework augmented by a knowledge distillation module that guides the classification model by the WSD model according to the class-level predictions for the whole image and the object-level visual features for object RoIs.
The WSD model is the \emph{teacher} model and the classification model is the \emph{student} model.
After this cross-task knowledge distillation, the performance of the classification model is significantly improved and the efficiency is maintained since the WSD model can be safely discarded in the test phase.
Extensive experiments on two large-scale datasets (MS-COCO and NUS-WIDE) show that our framework achieves superior performances over the state-of-the-art methods on both performance and efficiency.
\end{abstract}

%
%


\copyrightyear{2018}
\acmYear{2018}
\setcopyright{acmcopyright}
\acmConference[MM '18]{2018 ACM Multimedia Conference}{October 22--26, 2018}{Seoul, Republic of Korea}
\acmBooktitle{2018 ACM Multimedia Conference (MM '18), October 22--26, 2018, Seoul, Republic of Korea}
\acmPrice{15.00}

\maketitle


\section{Introduction}
Multi-label image classification (MLIC) \cite{MLIC27_ijdwm09_tgit,MLIC25_ijca17_RS} is one of the pivotal and long-lasting problems in computer vision and multimedia.
This task starts from the observation that real-world images always contain diverse semantic contents that need multiple visual concepts to classify.
Except for the challenges shared with single-label image classification (e.g., inter-class similarity and intra-class variation), MLIC is more difficult because predicting the presence of multiple classes usually needs a more thorough understanding of the input image (e.g., associating classes with semantic regions and capturing the semantic dependencies of classes).

\begin{figure}[t]
\centerline{\includegraphics[width=8.4cm]{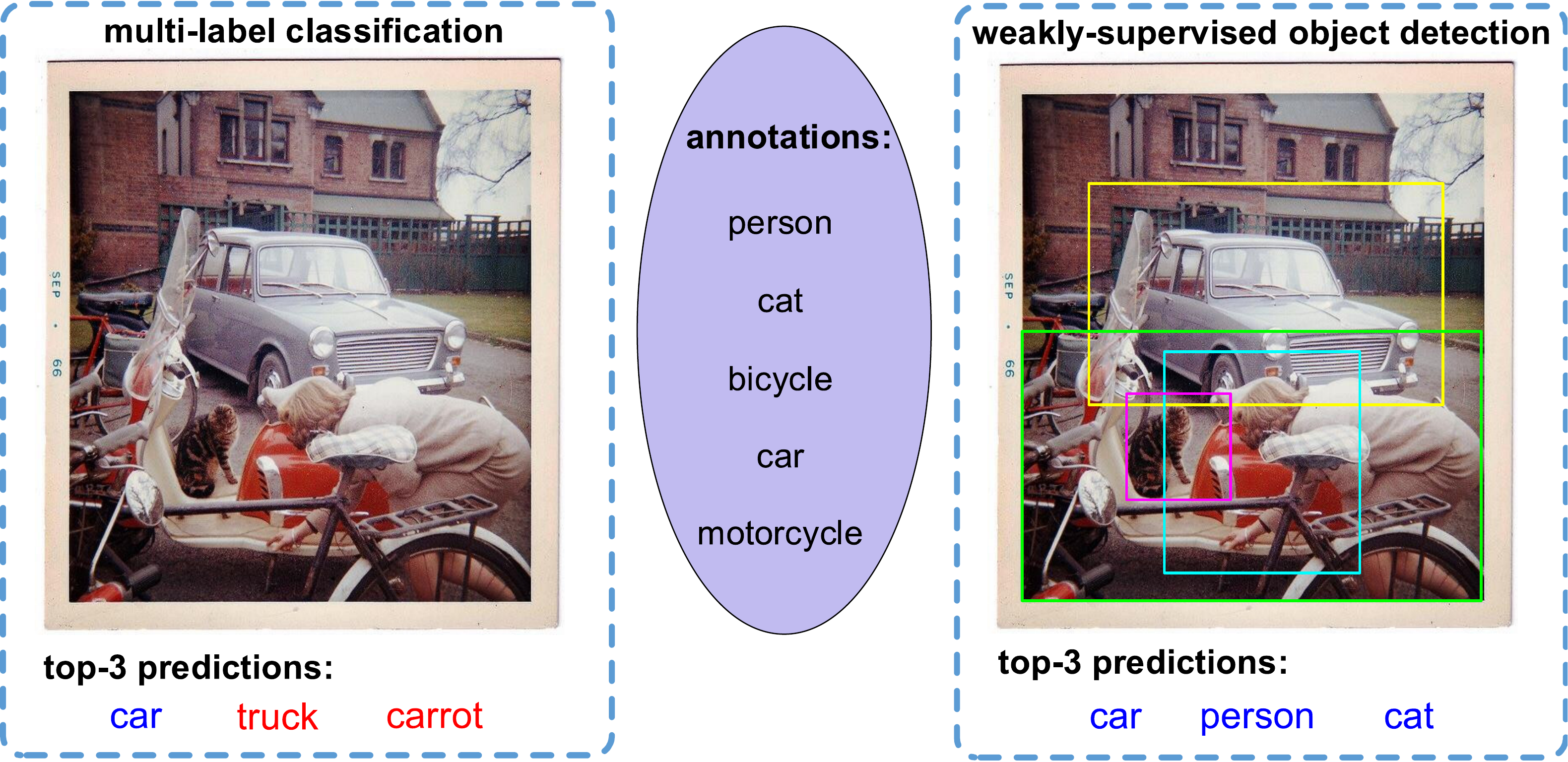}}
\vspace{0pt}\caption{\small The illustration of multi-label image classification (MLIC) and weakly-supervised detection (WSD).
We show top-3 predictions, in which correct predictions are shown in \textcolor[rgb]{0.00,0.00,1.00}{blue} and incorrect predictions in \textcolor[rgb]{1.00,0.00,0.00}{red}.
The MLIC model might not predict well due to poor localization for semantic instances.
Although the detection results of WSD may not preserve object boundaries well, they tend to locate the semantic regions which are informative for classifying the target object, such that the predictions can still be improved.}
\label{Fig:introduction}
\vspace{-10pt}
\end{figure}

Contemporary methods may simply finetune the multi-label classification networks pre-trained on the single-label classification datasets (e.g., ImageNet \cite{ImageNet_ijcv15_ojhjsszaa}).
However, the classifiers trained for global image representations may not generalize well to the images in which objects from multiple classes are distributed in different locations, scales and occlusions.
To mitigate this problem, the task of MLIC can be decomposed into multiple independent binary classification tasks, in which one classifier only focuses on one object label.
In this way, though very efficient, the semantic dependencies among multiple classes, which is especially important for MLIC~\cite{MLIC6_cvpr16_JYJZCW}, are ignored (e.g., ``cat'' is more likely to be misclassified into the category of ``dog'' than falsely associated to ``car'').
Therefore, some prior works~\cite{MLIC6_cvpr16_JYJZCW,MLIC14_eccv16_hjj,MLIC18_aaai18_SYCY} tried to fix this drawback by explicitly capturing the class dependencies with a RNN or LSTM structure appended after CNN-based models.
However, they usually suffer from the difficulty in back-propagating stable gradients \cite{MLIC3_cvpr17_FTTWC}.


Recently, some object localization techniques~\cite{MLIC10_pami16_ywjbjys,MLIC21_arxiv17_JQJCJ, MLIC22_tmm18_jqjcj} are introduced into the MLIC task by simplifying the multi-label classification problem into multi-object detection task.
The resulting pipeline usually involves two steps.
The hypothesis regions are first proposed using low-level image cues~\cite{WSD4_iccv11_kjta}.
Then a neural network is trained to predict class scores on these proposals, and these predictions are aggregated to achieve MLIC task.
Even though satisfactory performance can be achieved with sufficiently accurate region proposal algorithms, these methods always have to bear redundant computational cost in the test phase. Thus they are usually not practical for large-scale applications.


To solve above issues, an effective and efficient multi-class image classification model needs to simultaneously hold three important advantages: (1) locating semantic regions for object-level feature extraction; (2) capturing semantic dependencies among multiple classes; (3) fewer additional computation and annotation budgets for the practical issue.
Following this intuition, weakly-supervised detection (WSD)~\cite{WSD2_cvpr16_ha} may be a feasible solution.
It could achieve the detection goal of locating each semantic instance with a specific class using only image-level annotations.
Figure~\ref{Fig:introduction} shows the task illustrations for MLIC and WSD frameworks.
The MLIC model might not predict well due to the lack of object-level feature extraction and localization for semantic instances.
Although the results detected by WSD may not preserve object boundaries well, they tend to locate the semantic regions which are informative for classifying the target object, such that the predictions can still be improved.
Therefore, the localization results of WSD could provide object-relevant semantic regions while its image-level predictions could naturally capture the class dependencies.
These unique advantages are very useful for the MLIC task.
The only problem is the huge computational complexity in the WSD pipelines.
Is it possible to combine the advantages in WSD with the high efficiency of simple classification network?
Knowledge distillation~\cite{KD1_nips15_goj}, a technique that distills knowledge from a large \emph{teacher} model into a much smaller \emph{student} model, may provide a good solution to guide the classification model to inherently contain object-level localization ability and mutual class dependencies.


\begin{figure*}[t]
\centerline{\includegraphics[width=16.5cm]{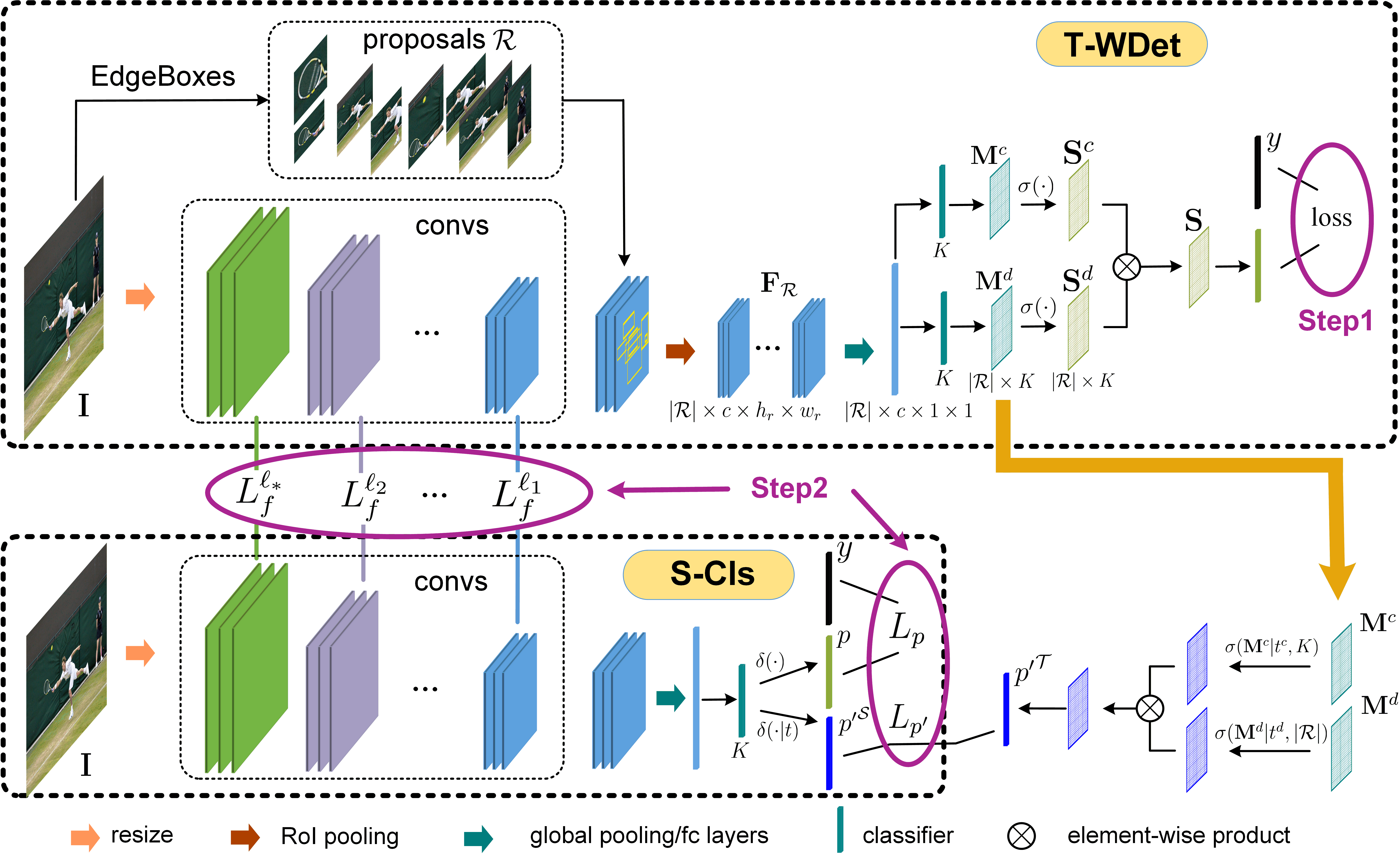}}
\vspace{-11pt}\caption{\small The overall architecture of our framework. The proposed framework works with two steps: (1) we first develop a WSD model as teacher model (called T-WDet) with only image-level annotations $y$; (2) then the knowledge in T-WDet is distilled into the MLIC student model (called S-Cls) via feature-level distillation from RoIs and prediction-level distillation from whole image, where the former is conducted by optimizing the loss $\sum_{\ell}^{}{\bm L}_f^{\ell}$ while the latter is conducted by optimizing the loss ${\bm L}_{p}$ and ${\bm L}_{p'}$.}
\label{Fig:overall_architecture}
\vspace{-11pt}
\end{figure*}

In this paper, we propose a novel and efficient deep framework to boost MLIC by distilling the unique knowledge from WSD into classification with only image-level annotations.
The overall architecture of our framework is illustrated in Figure~\ref{Fig:overall_architecture}.
Specifically, our framework works with two steps: (1) we first develop a WSD model with image-level annotations; (2) then we construct an end-to-end knowledge distillation framework by propagating the class-level holistic predictions and the object-level features from RoIs in the WSD model to the MLIC model, where the WSD model is taken as the \emph{teacher} model (called T-WDet) and the classification model is the \emph{student} model (called S-Cls).
The distillation of object-level features from RoIs focuses on perceiving localizations of semantic regions detected by the WSD model while the distillation of class-level holistic predictions aims at capturing class dependencies predicted by the WSD model.
After this distillation, the classification model could be significantly improved and no longer need the WSD model, thus resulting in high efficiency in test phase.


The main contributions of this work are highlighted as follows:
\vspace{5pt}
\begin{itemize}
\setlength{\itemsep}{0ex}
\item A novel and efficient deep multi-label image classification framework equipped by knowledge distillation is proposed, which distills the unique knowledge from a weakly-supervised detection model into the classification model such that the latter is improved significantly with high efficiency.
\item To our best knowledge, it is the first work that applies knowledge distillation between two different tasks, i.e., weakly-supervised detection and multi-label image classification.
\item Extensive experiments are conducted on two large-scale public datasets (MS-COCO and NUS-WIDE), and the results show that our framework achieves superior performances over the state-of-the-art methods on both performance and efficiency.
\end{itemize}
\vspace{-13pt}
\section{Related work}

\noindent \textbf{Multi-Label Image Classification (MLIC).} \ The progress of MLIC \cite{MLIC4_iclr14_YYTAS,MLIC10_pami16_ywjbjys,MLIC20_pr17_PXZXW,MLIC5_cvpr17_FHWNX} has been greatly made with deep convolutional neural network \cite{AlexNet_nips2012_aig,VGG_iclr15_ka,ResNet_cvpr16_hzrs}.  Some works \cite{MLIC1_iccv15_JLL,MLIC2_cvpr17_YYJ,MLIC3_cvpr17_FTTWC,MLIC6_cvpr16_JYJZCW,MLIC8_cvpr16_QMWD,MLIC9_cvpr15_MQACJ,MLIC13_aaai17_CWWYF,MLIC14_eccv16_hjj,MLIC16_iccv17_PRLE,MLIC21_arxiv17_JQJCJ,MLIC24_cvpr16_hgzzg}  embed label dependencies with the deep model to improve the accuracy of MLIC. CNN-RNN \cite{MLIC6_cvpr16_JYJZCW} utilizes RNN combined with CNN to learn a joint image-label embedding for capturing label dependencies. \cite{MLIC3_cvpr17_FTTWC} proposes a regularised embedding layer as the interface between the CNN and RNN to mitigate the difficulty of model training in \cite{MLIC6_cvpr16_JYJZCW}. \cite{MLIC8_cvpr16_QMWD,MLIC9_cvpr15_MQACJ,MLIC26_mm17_javpzh} learn graph structure to model the label dependencies. These methods always require pre-defined label relations. Some other works ensemble multiple deep models with different input scales \cite{MLIC17_aaai18_TZGL,MLIC7_iccv17_ZTGRL} while suffering high complexity.

Recently, various methods \cite{MLIC5_cvpr17_FHWNX,MLIC7_iccv17_ZTGRL,MLIC11_cvpr17_hjjy,MLIC12_cvpr16_hjybjj,MLIC17_aaai18_TZGL,MLIC18_aaai18_SYCY,MLIC22_tmm18_jqjcj} have been proposed to locate semantic regions for learning deep attentional representations. For example, MIML-FCN+ \cite{MLIC11_cvpr17_hjjy} uses bounding boxes from Faster-RCNN \cite{Faster-RCNN_nips15_skfj} to locate the objects in an image for multi-instance learning. Spatial regularization network \cite{MLIC5_cvpr17_FHWNX} generates class-related attention maps to capture
spatial dependencies. \cite{MLIC7_iccv17_ZTGRL} employs a LSTM sub-network to predict labeling scores on the regions located by a spatial transformer layer.

All the aforementioned methods either require pre-defined label relations or object-level annotations, and they usually add model complexity by the extra modules, both of which result in poor practicality.

\noindent \textbf{Weakly-supervised detection (WSD).} \ Recently, many researches on WSD \cite{WSD1_mm17_xdfy,WSD2_cvpr16_ha,WSD5_cvpr17_djysm,WSD6_iccv17_ky,WSD7_iccv17_yyqqj} have been conducted. Dual-network \cite{WSD1_mm17_xdfy} is proposed to optimize proposal generation and instance selection in a joint framework. \cite{WSD5_cvpr17_djysm} introduces the domain adaptation techniques for the WSD task.
WSDDN \cite{WSD2_cvpr16_ha} modifies ImageNet pre-trained VGG \cite{VGG_iclr15_ka} to operate at image regions, performing simultaneously region selection and classification. In this work, we extend the WSDDN to develop a WSD model into our framework.

\noindent \textbf{Knowledge Distillation.} \ Hinton et al. \cite{KD1_nips15_goj} use a softened version of the output of a large \emph{teacher} network to teach information to a small \emph{student} network. FitNets \cite{KD2_iclr15_ansacy} employs not only the output but also intermediate layer values of the teacher network to train the student network. Attention transfer \cite{KD3_iclr17_sn} forces the student network to be consistent with the teacher network on feature attention maps. These methods focus on the distillation between the same tasks, and they always use the whole feature maps and class-identical soften targets to conduct distillation, which can not locate to semantic regions of the image and are not sensitive to classes. Chen et al. \cite{Chen_nips2017_gwxtm} concentrate on distilling between the same tasks of object detection while our proposed distillation is operated between two different tasks, i.e., from WSD to MLIC.

\section{Methodology}
Multi-label image classification aims at obtaining all the semantic classes in an image. Generally, given an image $\mathbf{I}$, the final prediction $l_k$ of the $k$-th class corresponding to $\mathbf{I}$ is formulated by
\begin{equation}
\label{Eq:problem_formula}
l_k = \mathds{I}(p_k(\mathbf{I}|{\mathbf w}) >  {\tau}_k), \ k\in\{1,\cdots,K\},
\end{equation}
where $p_k(\mathbf{I}|\mathbf{w})$, estimated by a model with parameters $\mathbf{w}$, denotes the posterior probability of image $\mathbf{I}$ including the $k$-th class. $K$ is the number of given labels, and ${\tau}_k$ is the confidence threshold for the $k$-th class. $\mathds{I}(p > \tau)$ is an indicator function, it takes $1$ when $p > \tau$ and $0$ otherwise. $l_k$ is the final label indicator, i.e., $l_k = 1$ means the $k$-th class is included in the given image and $l_k = 0$ otherwise.

In this paper, we propose a novel and efficient framework in which the multi-label image classification (MLIC) task is facilitated by weakly-supervised detection task. In the following, we will present the proposed framework in detail, including (1) weakly-supervised detection (WSD) model, (2) Knowledge distillation from WSD to MLIC, and (3) Implementation details.

\subsection{Weakly-Supervised Detection (WSD) Model}
\label{Sec:WSD}

Although any existing WSD methods can be used in our framework, we choose WSDDN \cite{WSD2_cvpr16_ha} because of its architecture accessibility.
Using VGG16  \cite{VGG_iclr15_ka} pre-trained on ImageNet \cite{ImageNet_ijcv15_ojhjsszaa} as backbone network, WSDDN operates on image regions which are outputted by EdgeBoxes (EB) \cite{WSD3_eccv14_cp}.
In this work, we extend WSDDN to support any popular networks.
The architecture of extended WSDDN (called T-WDet) is illustrated at the upper part of Figure \ref{Fig:overall_architecture}.
First, EB algorithm is used to get a lot of proposals $\mathcal{R}$ from the input image $\mathbf{I}$.
These proposals are inputted to the RoI pooling \cite{Faster-RCNN_nips15_skfj} module to get RoI-localized features.
Note that we replace SPP pooling \cite{SPP_pami15_hxsj} by RoI pooing, because the latter keeps the spatial information.
Formally, let $\mathbf{F}_{\rm conv} \in \mathbb{R}^{c\times h\times w}$ denote the last convolutional feature maps of the backbone network, $R$ denote a proposal in $\mathcal{R}$, and $s_R$ denote the prior score of $R$ outputted by EB algorithm, $\mathbf{F}_{\mathcal{R}} \in \mathbb{R}^{\left| \mathcal{R} \right| \times c\times h_r\times w_r}$, the obtained RoI features of all the proposals,  can be described as
\begin{equation}
\label{Eq:roi_pooling}
\begin{split}
& \mathbf{F}_{R} = s_R \odot \phi_{\rm RoI}(\mathbf{F}_{\rm conv} ; R),  \\
& \mathbf{F}_{\mathcal{R}} = {\mathds{C}}_{R \in \mathcal{R}} (\mathbf{F}_{R}),
\end{split}
\end{equation}
where $\phi_{\rm RoI}(\cdot)$ is the operation of RoI pooling, ``$\odot$'' is the operation of multiplying each element of $\phi_{\rm RoI}(\cdot)$ by score $s_R$, and $\mathds{C}(\cdot)$ is the concatenation operation, which concatenates the features of $\left| \mathcal{R} \right|$ proposals along the fourth dimension.

Then, global pooling or several fully connected (fc) layers are adopted to further transform the RoI feature maps $\mathbf{F}_{\mathcal{R}}$ into feature vectors. The subsequent network is split into two branches. Both of them pass through a fully connected layer, where the output is consistent with the given classes $K$, to get a logit matrix $\mathbf{M} \in \mathbb{R}^{\left| \mathcal{R} \right| \times K}$. One branch aims at classification while the other at detection. The classification is achieved by a softmax operation along the first dimension $K$, and the detection along the second dimension $\left| \mathcal{R} \right|$.

Finally,  the element-wise product operation is adopted to fuse the softmax score matrix $\mathbf{S^c},\mathbf{S^d} \in \mathbb{R}^{\left| \mathcal{R} \right| \times K}$ of the two branches. The fused score matrix $\mathbf{S}$ is summed along the second dimension $\left| \mathcal{R} \right|$ to get final class prediction $p \in \mathbb{R}^{K}$, which is compatible with the image-level annotations $y \in \mathbb{R}^{K}$. The final detection results for each class are obtained by processing each column of $\mathbf{S}$ with non-maximum suppression (NMS). T-WDet is also trained in an end-to-end manner. More details can be referred in \cite{WSD2_cvpr16_ha}.

\subsection{Knowledge Distillation from WSD to MLIC}

In this paper, we argue that there is unique knowledge beyond classification contained in the task of WSD, which could facilitate MLIC.
Specifically, on one hand, the detection results of WSD provide localization of semantic regions, which is a powerful cue for classification model to further understand the image.
On the other hand, the image-level prediction confidences of WSD naturally capture semantic dependencies among classes, which could be a strong reference for MLIC from the perspective of detection.

As stated in Section \ref{Sec:WSD}, we first use a T-WDet model to achieve the goal of detection. The problem locates at how to transfer the unique knowledge from T-WDet model into the classification model. A reasonable solution is knowledge distillation \cite{KD1_nips15_goj}, which can distill the knowledge in a large \emph{teacher} model for improving a small \emph{student} model. Inspired by this idea, we propose a dedicated distillation framework to distill knowledge from a WSD \emph{teacher} model (T-WDet) for boosting a MLIC \emph{student} model (S-Cls). This distillation framework works with two stages. The first stage focuses on the feature-level knowledge transfer while the second stage on the prediction-level knowledge transfer. Both of the two stages are included in ``step 2'' in Figure \ref{Fig:overall_architecture}.

\vspace{+1mm}
\noindent \textbf{Feature-level knowledge transfer.} \ We propose a RoI-aware distillation approach which explicitly distills the localization knowledge from WSD to MLIC at feature level. Specifically, we sum the fused score matrix $\mathbf{S}$ in the well-trained T-WDet model along the first dimension $K$ to get a confidence vector $s' \in \mathbb{R}^{\left| \mathcal{R} \right|}$.
This vector implies a confidence distribution of all the proposals' objectness, i.e., region proposal score, which is a reliable localization importance indicator for MLIC.
Since the proposals outputted by EB algorithm are highly overlapped, we take NMS operation for them using the obtained confidences $s'$. Then, with these well-chosen proposals, the knowledge from T-WDet model is distilled into S-Cls model by minimizing the $\ell 2$ loss of RoI pooled features on selected convolutional layers as
\begin{equation}
\label{Eq:loss_featureLevel}
{\bm L}_f ({\mathbf w}_{\rm conv}^{\mathcal{S}}) = \frac{1}{2N} \sum\nolimits_{n} \frac{1}{\left| \mathcal{R}_n' \right|}
\lVert \mathbf{F}^{\mathcal{T}}_{\mathcal{R}_n'} \ominus \mathbf{F}^{\mathcal{S}}_{\mathcal{R}_n'} \rVert_2^2,
\end{equation}
where $\mathcal{R}_n'$ denotes the remaining proposals after performing NMS to $\mathcal{R}_n$ for image $\mathbf{I}_n$ and $N$ is the number of training images. ``$\ominus$'' is the element-wise subtraction operation. $\mathbf{F}^{\mathcal{T}}_{\mathcal{R}_n'}$ and $\mathbf{F}^{\mathcal{S}}_{\mathcal{R}_n'}$ denote the RoI pooled features from T-WDet model and S-Cls model, respectively. They can be described as
\begin{equation}
\label{Eq:roi_feature}
\begin{split}
& \mathbf{F}^{\mathcal{T}}_{\mathcal{R}_n'} =  {\mathds{C}}_{R \in \mathcal{R}_n'} \big[ s'_{R}  \odot  \phi_{\rm RoI}(\mathbf{F}^{\mathcal{T}}_{\rm conv} ; R) \big], \\
& \mathbf{F}^{\mathcal{S}}_{\mathcal{R}_n'} =  {\mathds{C}}_{R \in \mathcal{R}_n'} \big[ s'_{R}  \odot  \phi_{\rm RoI}(\Psi(\mathbf{F}^{\mathcal{S}}_{\rm conv}) | {\mathbf w}_{\rm conv}^{\mathcal{S}}; R) \big], \\
\end{split}
\end{equation}
where $\mathbf{F}^{\mathcal{T}}_{\rm conv}$ and $\mathbf{F}^{\mathcal{S}}_{\rm conv}$ denote the selected convolutional layers in T-WDet model and S-Cls model, respectively. $\Psi(\mathbf{F}^{\mathcal{S}}_{\rm conv})$ is the possibly needed transforming operation, which transforms $\mathbf{F}^{\mathcal{S}}_{\rm conv}$ to be compatible with $\mathbf{F}^{\mathcal{T}}_{\rm conv}$ in case the number of their channels is different. In this process, we only update the convolutional parameters ${\mathbf w}_{\rm conv}^{\mathcal{S}}$ in S-Cls model and $s'_{R}$ plays a role as local importance weighting factor for proposal $R$.


Our RoI-aware distillation approach explicitly distills the unique knowledge from the detection results of T-WDet model into S-Cls model, i.e., localization of semantic regions and objectness confidence. It is superior to FitNets \cite{KD2_iclr15_ansacy} and attention transfer \cite{KD3_iclr17_sn}, because both of them transfer knowledge on whole feature map, which is not sensitive to localization and objectness. Our distillation approach can also be operated on multiple layers, then the loss we minimize becomes $\sum_{\ell}^{}{\bm L}_f^{\ell} ({\mathbf w}_{\rm conv}^{\mathcal{S}})$.

\vspace{+1mm}
\noindent \textbf{Prediction-level knowledge transfer.} \ The final label prediction $p$ of T-WDet model is obtained by summing the score matrix $\mathbf{S}$ along the second dimension $\left| \mathcal{R} \right|$. It aggregates the confidence of all the proposals over the given classes, which is a powerful reference for classification. Moreover, we observe that the classification accuracy for different classes between T-WDet model and S-Cls model are very different, thus the prediction-knowledge transfer should be of difference over classes. To discriminatively distill the knowledge from T-WDet model to S-Cls model at prediction level, we propose a class-aware distillation approach. Specifically, after initializing the parameters $\mathbf{w}^\mathcal{S}$ of S-Cls model with ${\mathbf w}_{\rm conv}^{\mathcal{S}}$ pre-trained in the first stage, we then simultaneously minimize two different loss functions for S-Cls model in this stage. The first loss function is the $\ell 2$ loss of the discriminatively softened predictions of T-WDet model and S-Cls model as
\begin{equation}
\label{Eq:loss_soft}
{\bm L}_{p'}({\mathbf w}^{\mathcal{S}}) =  \frac{1}{2N} \sum\nolimits_{n} \lVert {p'}^{\mathcal{T}} - {p'}^{\mathcal{S}}({\mathbf w}^{\mathcal{S}}) \rVert_2^2,
\end{equation}
where ${p'}^{\mathcal{T}}$ is the softened predictions of T-WDet model, which is calculated by
\begin{equation}
\label{Eq:soft_pred_tea}
{p'}^{\mathcal{T}} =  \sum_{i=1}^{\left| \mathcal{R} \right|} [ \sigma(\mathbf{M}^c|t^c, K) \otimes \sigma(\mathbf{M}^d|t^d, {\left| \mathcal{R} \right|}) ], \ \mathbf{M} \in \mathbb{R}^{\left| \mathcal{R} \right| \times K}.
\end{equation}
Here,  ``$\otimes$'' is the the element-wise product operation. $\sigma(\mathbf{M}^c|t^c, K)$ and $\sigma(\mathbf{M}^d|t^d, {\left| \mathcal{R} \right|})$ are the softened softmax operation along the first dimension $K$ (classification branch) and the second dimension $\mathcal{R}$ (detection branch) on the logit matrix $\mathbf{M}$, respectively. They can be defined as
\begin{equation}
\label{Eq:soft_softmax_tea}
\begin{split}
[\sigma(\mathbf{M}^c|t^c, K)]_{ij} &= \frac{e^{m^c_{ij}/t^c_k}}{\sum_{k=1}^{K}e^{m^c_{ik}/t^c_k}}, \ \forall \, i \in \{1, \cdots, {\left| \mathcal{R} \right|}\},   \\
[\sigma(\mathbf{M}^d|t^d, {\left| \mathcal{R} \right|})]_{ij} &= \frac{e^{m^d_{ij}/t^d_r}}{\sum_{r=1}^{{\left| \mathcal{R} \right|}}e^{m^d_{rj}/t^d_r}}, \ \forall \, j \in \{1, \cdots, K\},
\end{split}
\end{equation}
where $t^c_k$ and $t^d_r$ are the softmax temperature of $k$-th class and the softmax temperature of $r$-th proposal, respectively.

${p'}^{\mathcal{S}}({\mathbf w}^{\mathcal{S}})$  is the softened sigmoid predictions of S-Cls model, which is calculated by
\begin{equation}
\label{Eq:soft_pred_stu}
{p'}_k^{\mathcal{S}}({\mathbf w}^{\mathcal{S}}) = \delta(m_k|t) = 1/(1+e^{-m_k/t_k}),
\end{equation}
where $m_k$ and $t_k$ are the logit and the sigmoid temperature of $k$-th class, respectively. We decompose multi-label classification task as multiple binary classification tasks, and we use sigmoid operation to get the final output.

Note that all the temperatures are different, and they are learnable in the training phase. This is more reasonable for our task than the class-identical and fixed temperature used in \cite{KD1_nips15_goj}. Moreover, it also cuts down the laborious costs for tuning the artificial temperatures by this learnable way. Formally, let $m$ denote the input data, $t$ denote the temperature and $\hat{m}$ denote the output data: $\hat{m}_i = m_i/t_i$ , then the back-propagation and chain rule are used to compute derivatives w.r.t $m$ and $t$ as
\begin{equation}
\label{Eq:deriv_temp}
\frac{\partial {\bm L}_{p'}}{\partial m_i} = \sum\nolimits_{t_i} \frac{\partial {\bm L}_{p'}}{\partial \hat{m}_i} \frac{1}{t_i}, \ \
\frac{\partial {\bm L}_{p'}}{\partial t_i} = \sum\nolimits_{m_i} \frac{\partial {\bm L}_{p'}}{\partial \hat{m}_i} (-\frac{m_i}{t_i^2}).
\end{equation}

The second loss function is the cross entropy with hard label (ground truth) $y$ as
\begin{equation}
\label{Eq:loss_hard}
{\bm L}_{p}({\mathbf w}^{\mathcal{S}}) = - \frac{1}{N} \sum\nolimits_{n} [y\,\text{log}\, p + (1-y)\text{log}(1-p)],
\end{equation}
where $p$ is the normal sigmoid prediction of S-Cls model.

In the class-aware distillation stage, we update all the parameters $\mathbf{w}^{\mathcal{S}}$ of S-Cls model. For one thing, the S-Cls model fits the given hard labels by working as multiple binary classification tasks. For another, it also acquires the knowledge of detection, i.e., semantic dependencies of classes distilled from well-trained T-WDet model.
{\renewcommand\baselinestretch{1}\selectfont
\renewcommand{\algorithmicrequire}{\textbf{Input:}}   
\renewcommand{\algorithmicensure}{\textbf{Output:}}
\renewcommand{\algorithmicrepeat}{\textbf{Repeat:}}   
\renewcommand{\algorithmicuntil}{\textbf{Until:}}
\renewcommand{\algorithmicreturn}{\textbf{Return:}}
\begin{algorithm}[t]
\caption{\ Training and Test of S-Cls model}
\label{alg:train_test}
\begin{algorithmic}[1]
\TRAINING ~~ \\
\REQUIRE image data and label data $(\mathbf{I}^N,y^N)$.
\ENSURE parameters $\mathbf{w}$ of S-Cls model.
\INITIALIZE $\mathbf{w}$, $\lambda$ and training hyper-parameters.
\STAGEONE Feature-Level Knowledge Transfer.
\REPEAT
\STATE compute ${\bm L}_f ({\mathbf w}_{\rm conv}^{\mathcal{S}})$ by Eq. (\ref{Eq:loss_featureLevel}), Eq. (\ref{Eq:roi_feature}).
\STATE update ${\mathbf w}_{\rm conv}^{\mathcal{S}}$ by gradient back-propagation.
\UNTIL ${\bm L}_f ({\mathbf w}_{\rm conv}^{\mathcal{S}})$ converges.
\STAGETWO Prediction-Level Knowledge Transfer.
\REPEAT
\STATE compute ${\bm L}_{p}({\mathbf w}^{\mathcal{S}}) + {\lambda} {\bm L}_{p'}({\mathbf w}^{\mathcal{S}})$ by Eq. (\ref{Eq:loss_soft}), Eq. (\ref{Eq:loss_hard}), Eq. (\ref{Eq:loss_all}).
\STATE update ${\mathbf w}^{\mathcal{S}}$ by gradient back-propagation.
\UNTIL ${\bm L}_{p}({\mathbf w}^{\mathcal{S}})$ converges.
\MYRETURN ${\mathbf w}^{\mathcal{S}}$
\vspace{+1mm}
\TESTING ~~ \\
\REQUIRE image data $\mathbf{I}^N$.
\ENSURE prediction $l^N$.
\INITIALIZE parameters $\mathbf{w}$ of S-Cls model, confidence threshold $\tau$.
\FOR {$n=1$ to $N$}
\STATE forward pass S-Cls model to get $p(\mathbf{I}^n|{\mathbf w})$.
\STATE compute $l^n$ by Eq. (\ref{Eq:problem_formula}).
\ENDFOR
\MYRETURN $l^N$
\end{algorithmic}
\end{algorithm}
\par} 

\subsection{Implementation Details}
\noindent \textbf{Training.} \ In the training phase, we first train a T-WDet model as stated in Section \ref{Sec:WSD}. Then, we froze all the parameters of well-trained T-WDet model, and train the S-Cls model using the proposed RoI-aware and class-aware distillation framework. This is operated with two stages. \textbf{Stage 1:} We train S-Cls model to update convolutional parameters ${\mathbf w}_{\rm conv}^{\mathcal{S}}$ by optimizing the loss in Eq. (\ref{Eq:loss_featureLevel}). \textbf{Stage 2:} We update the whole network by optimizing the weighted losses in Eq. (\ref{Eq:loss_hard}) and Eq. (\ref{Eq:loss_soft}) as
\begin{equation}
\label{Eq:loss_all}
{\bm L}_{p}({\mathbf w}^{\mathcal{S}}) + {\lambda} {\bm L}_{p'}({\mathbf w}^{\mathcal{S}}),
\end{equation}
where ${\lambda}$ is the weighted factor.

\noindent \textbf{Test.} \ In the test phase, the S-Cls model works without T-WDet model. It is compact as the same as standard classification model, i.e., no any extra computational cost. The normal sigmoid outputs $p$ is taken as its final predictions. The pseudo-code of training and test of S-Cls model can be referred in Algorithm \ref{alg:train_test}.

To convincingly demonstrate the proposed framework, we use VGG16 pre-trained on ImageNet as backbone network for both T-WDet and S-Cls models. VGG16 is the most popular network used in the literature of MLIC, thus a fair comparison can be made. The RoI pooling size is set to $7\times7$ for the two networks. The image size input to S-Cls model is always $224\times224$.

\noindent \textbf{T-WDet model.} \ For training with mini-batch, we take top $500$ proposals of EB algorithm, which are sorted by the prior scores of EB. Moreover, we recycle high-score proposals if the candidate number is less than $500$. Except mirror flip, we train T-WDet model also using the popular techniques applied in detection area, i.e., training with multiple square scales of $480$, $576$, $672$, $736$ and $832$. We use the input scale of $672$ in the distillation process.

\noindent \textbf{Knowledge  distillation.} \ In feature-level knowledge transfer, the NMS threshold is set to $0.4$ to clean highly overlapped proposals. Moreover, we take top $100$ proposals after NMS for training with mini-batch, and again, we recycle them if the candidate number is not enough. In this stage, we only use the \texttt{conv5\_3} layer for knowledge transfer. The transforming operation is set to $\Psi(\mathbf{F}) = \mathbf{F}$ due to the equal number of channels between two networks. In prediction-level knowledge transfer, the convolutional layers of S-Cls model are initialized with ${\mathbf w}_{\rm conv}^{\mathcal{S}}$ trained in last stage, fully connected layers are initialized with ImageNet pre-trained parameters, and other layers are initialized with Xavier algorithm \cite{Xavier_jmlr10_gb}. The value of all the temperatures $t$ are initialized with $1$. The weighted factor ${\lambda}$ of two losses is set to $1$.

Our framework is implemented using Caffe \cite{Caffe_mm14_jskl}. The stochastic gradient descend (SGD) algorithm is employed for the network training, with a batchsize of $32$, momentum of $0.9$ and weight decay of $0.0005$. For feature-level transfer, the learning rate is fixed at $10^{-5}$ and the training continues for about $100$ epochs. For prediction-level transfer, the initial learning rate is set to $10^{-4}$, and decreased to $1/10$ when validation loss gets saturated.
\begin{figure*}[t]
\centerline{\includegraphics[width=17.5cm]{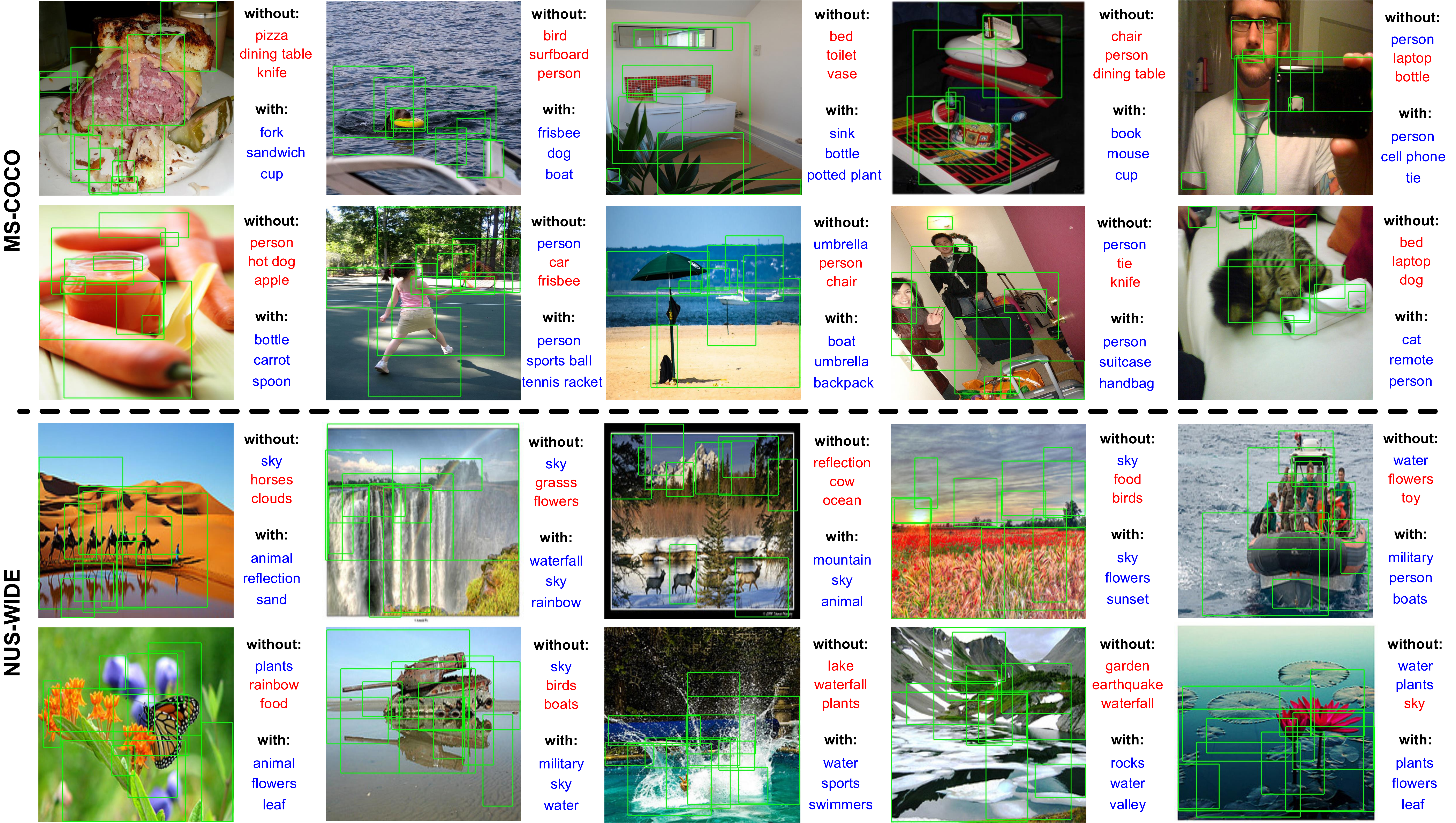}}
\vspace{-5pt}
\caption{\small Example results on two datasets. The green bounding boxes in images are the top-10 proposals detected by T-WDet model, which is sorted by objectness confidences $s'$ in Eq. \ref{Eq:roi_feature}. The text on the right of images are the top-3 classification results of S-Cls model ``without'' and ``with'' knowledge distillation using our framework, where correct predictions are shown in \textcolor[rgb]{0.00,0.00,1.00}{blue} and incorrect predictions in \textcolor[rgb]{1.00,0.00,0.00}{red}.}
\label{Fig:results_show}
\vspace{-5pt}
\end{figure*}
\section{Experiment}
\subsection{Datasets}
The proposed framework is evaluated on two large-scale datasets with fairly different types of labels: MS-COCO \cite{coco_eccv14_ty} with 80 object labels and NUS-WIDE \cite{nuswide_cvir09_cjn} with 81 concept labels.

\noindent \textbf{MS-COCO.} \ It contains $122$,$218$ images of 80 object labels, with about $2.9$ labels per image. The objects are of high diversity, and they are of severe occlusions. We follow the official split of $82$,$081$ images for training and $40$,$137$ validation images for testing.

\noindent \textbf{NUS-WIDE.} \ This dataset contains $269$,$648$ images and $5018$ tags from Flickr. There are a total of $1000$ tags after removing noisy and rare tags. These images are further manually annotated into 81 concepts with $2.4$ concepts per image on average. The concepts are quite diverse, including event (e.g., running), scene/location (e.g., airport), object (e.g., animal). We follow the split used in \cite{MLIC4_iclr14_YYTAS,MLIC3_cvpr17_FTTWC}, i.e., $150$,$000$ images for training and $59$,$347$ for testing after removing the images without any labels.

Note that both of the two datasets are imbalanced over classes, and the imbalance on NUS-WIDE is even worse.
\begin{table}[t]
  \centering
  \small
  \caption{\small Quantitative comparison (\%) on MS-COCO. ``w/'' and ``w/o'' indicate ``with'' and ``without'' knowledge distillation by the proposed framework, respectively. The values in bold are the best while the values underlined are the second best.}
  \vspace{-5pt}
  \label{tab:coco}
  \begin{tabular}{l|ccc|cc}
  \Xhline{1.2pt}
  \multirow{2}*{Method}&\multicolumn{3}{c|}{All}&\multicolumn{2}{c}{Top-3} \\
  \cline{2-6}
  &mAP&F1-C&F1-O&F1-C&F1-O \\
  \Xhline{1.2pt}
  CNN-RNN \cite{MLIC6_cvpr16_JYJZCW}&-&-&-&60.4&67.8 \\
  CNN-LSEP \cite{MLIC2_cvpr17_YYJ} &-&62.9&68.3&-&- \\
  CNN-SREL-RNN \cite{MLIC3_cvpr17_FTTWC}&-&63.4&\underline{72.5}&-&- \\
  RMAM($512$+$10$crop) \cite{MLIC7_iccv17_ZTGRL}&72.2&-&-&\underline{66.5}&\underline{71.3} \\
  RARLF($512$+$10$crop) \cite{MLIC17_aaai18_TZGL}&-&-&-&65.6&70.5 \\
  MIML-FCN-BB \cite{MLIC11_cvpr17_hjjy}&66.2&-&-&-&- \\
  MCG-CNN-LSTM \cite{MLIC22_tmm18_jqjcj}&64.4&-&-&58.1&61.3 \\
  RLSD \cite{MLIC22_tmm18_jqjcj}&68.2&-&-&62.0&66.5 \\
 \hline
  Ours-S-Cls (w/o)&70.9&63.6&67.0&60.7&66.7 \\
  Distillation \cite{KD1_nips15_goj}&71.3&64.7&69.3&61.5&67.6 \\
  FitNets \cite{KD2_iclr15_ansacy}&\underline{72.5}&\underline{65.2}&70.9&62.3&68.3 \\
  Attention transfer \cite{KD3_iclr17_sn}&71.4&64.6&69.8&61.6&67.8 \\
  \hline
  Ours-S-Cls (w/)&\textbf{74.6}&\textbf{69.2}&\textbf{74.0}&\textbf{66.8}&\textbf{72.7} \\
  \Xhline{1.2pt}
  \end{tabular}
\end{table}

\begin{table}[t]
  \centering
  \small
  \caption{\small Quantitative comparison (\%) on NUS-WIDE. ``w/'' and ``w/o'' indicate ``with'' and ``without'' knowledge distillation by the proposed framework, respectively. The values in bold are the best while the values underlined are the second best.}
  \vspace{-5pt}
  \label{tab:nus}
  \begin{tabular}{l|ccc|cc}
  \Xhline{1.2pt}
  \multirow{2}*{Method}&\multicolumn{3}{c|}{All}&\multicolumn{2}{c}{Top-3} \\
  \cline{2-6}
  &mAP&F1-C&F1-O&F1-C&F1-O \\
  \Xhline{1.2pt}
  CNN-RNN \cite{MLIC6_cvpr16_JYJZCW}&-&-&-&34.7&55.2 \\
  Tag-Neighbors \cite{MLIC1_iccv15_JLL}&52.8&-&-&45.2&62.5 \\
  CNN-LSEP \cite{MLIC2_cvpr17_YYJ}&-&52.9&70.8&-&- \\
  CNN-SREL-RNN \cite{MLIC3_cvpr17_FTTWC}&-&52.8&\underline{71.0}&-&- \\
  MCG-CNN-LSTM \cite{MLIC22_tmm18_jqjcj}&52.4&-&-&46.1&59.9 \\
  RLSD \cite{MLIC22_tmm18_jqjcj}&54.1&-&-&46.9&60.3 \\
  KCCA \cite{MLIC19_pr17_tlla}&52.2&-&-&-&- \\
  \hline
  Ours-S-Cls (w/o)&55.6&52.0&67.2&47.5&64.8 \\
  Distillation \cite{KD1_nips15_goj}&57.2&54.3&69.5&50.3&67.5 \\
  FitNets \cite{KD2_iclr15_ansacy}&57.4&54.9&70.4&51.4&68.6 \\
  Attention transfer\cite{KD3_iclr17_sn}&\underline{57.6}&\underline{55.2}&70.3&\underline{51.7}&\underline{68.8} \\
  \hline
  Ours-S-Cls (w/)&\textbf{60.1}&\textbf{58.7}&\textbf{73.7}&\textbf{53.8}&\textbf{71.1} \\
  \Xhline{1.2pt}
  \end{tabular}
\end{table}
\subsection{Evaluation Metrics and Compared Methods}

\noindent \textbf{Evaluation Metrics.}\ We employ three overall metrics for comparison: macro/micro F1 (``F1-C''/``F1-O'') and mean average precision (mAP). Macro F1 is evaluated by averaging per-class F1, while micro F1 is evaluated on the results of all the images over all classes. For computing F1, we tune a class-independent confidence threshold, i.e., if the confidence is greater than this threshold, the prediction is taken as positive. We also report top-3 F1 sorted by the prediction confidences. mAP is the mean average precision over classes. Generally, mAP is of more reference, because it directly measures ranking quality and does not require choosing the final predictions.

\noindent \textbf{Compared Methods.}\ We compare our framework against the following stage-of-the-art deep learning methods: \textbf{CNN-RNN} \cite{MLIC6_cvpr16_JYJZCW} and \textbf{CNN-SREL-RNN} \cite{MLIC3_cvpr17_FTTWC} employ CNN combined with RNN for classification; \textbf{CNN-LSEP} \cite{MLIC2_cvpr17_YYJ} estimates the optimal confidence thresholds for each class; \textbf{RMAM} \cite{MLIC7_iccv17_ZTGRL} and \textbf{RARLF} \cite{MLIC17_aaai18_TZGL} locate to image regions for classification, they use very large input size (512$\times$512) and multi-scale and multi-crop tricks in the test phase; \textbf{MIML-FCN-BB} \cite{MLIC11_cvpr17_hjjy} uses outputs from Faster RCNN \cite{Faster-RCNN_nips15_skfj} with bounding box annotations for classification; \textbf{RLSD} \cite{MLIC22_tmm18_jqjcj} employs RNN to capture dependencies at localized regions for classification; \textbf{Tag-Neighbors} \cite{MLIC1_iccv15_JLL} uses CNN to blend information from the image and its neighbors; \textbf{MCG-CNN-LSTM} \cite{MLIC22_tmm18_jqjcj} employs LSTM to capture dependencies at proposals of MCG \cite{MCG_cvpr14_djysm}. Note that for fair comparison, we only report the results of methods based on VGG16 network and results without using extra label information (e.g., detailed metadata in NUS-WIDE dataset) or ensemble testing (e.g., fusion of multi-scale and multi-crop test).

We also compare with three advanced distillation methods by implementing them following their paper: (1) Distillation \cite{KD1_nips15_goj}: the $t$ in T-WDet model and S-Cls model are tuned at $1$, $5$ and $2$, $5$ for MS-COCO and NUS-WIDE, respectively. (2) FitNets \cite{KD2_iclr15_ansacy}: we choose the middle layer \texttt{conv3\_3} as hint layer, then training with the setting in (1). (3) Attention transfer \cite{KD3_iclr17_sn}: as suggested in the paper, we choose \texttt{conv3\_3}, \texttt{conv4\_3} and \texttt{conv5\_3} as transfer layers, and the transfer is combined with (1).
\subsection{Experimental Results}

\noindent \textbf{MS-COCO.}\ Experimental results on this dataset are summarised in Table \ref{tab:coco}. With ImageNet pre-training, the simple S-Cls model can also perform well, and it outperforms the state-of-the-arts after knowledge distillation using our framework. Compared with RMAM \cite{MLIC7_iccv17_ZTGRL} and RARLF \cite{MLIC17_aaai18_TZGL}, which rely on a large input size and multi-crop test, our S-Cls model still performs better with small input (224$\times$224) and single-forward test. Moreover, our framework with only image-level annotations also outperforms those methods like MIML-FCN-BB, which require bounding box annotations. The framework also shows superior performance over other advanced distillation methods even though all of them get decent results. Some example results are shown at the upper part of Figure \ref{Fig:results_show}. As can be seen, although the proposals of T-WDet hold poor preserve of object boundaries, they locate at the semantic regions that are very informative for classification, thus the classification results are greatly improved. Taking the $1^{\text{st}}$ column of the $1^{\text{st}}$ row as an example, we can see the objects in the image are highly overlapped, resulting in a poor classification result, the top-3 predictions are all wrong. However, after distillation with detected semantic regions, even the occluded objects ``fork'' and ``cup'' can be well recognised.

\vspace{+1mm}
\noindent \textbf{NUS-WIDE.}\ Quantitative results on this dataset are summarised in Table \ref{tab:nus}. The simple S-Cls model again performs well with a backbone network pre-trained on ImageNet, and it also outperforms the state-of-the-arts after knowledge distillation with our framework. Meanwhile, compared with other architectures that add extra modules to the backbone network, e.g., CNN-RNN \cite{MLIC6_cvpr16_JYJZCW},  CNN-SREL-RNN \cite{MLIC3_cvpr17_FTTWC} and RLSD \cite{MLIC22_tmm18_jqjcj}, our framework performs better with higher efficiency at the same time. Moreover, the proposed framework also outperforms other advanced distillation methods. We also show some example results at the lower part of Figure \ref{Fig:results_show}. As it shows, the T-WDet model can still locate semantic regions even with the concept label, and the classification results are again improved by distilling these informative regions into the classification model. Taking the $2^{\text{nd}}$ column of the $1^{\text{st}}$ row as an example, the classification model only recognises correctly with one concept ``sky'' by a global perception of this image in top-3 predictions. However, other concepts like ``waterfall'' and ``rainbow'' are recognised after the distillation with our framework. This demonstrates the effectiveness and robustness of our framework simultaneously.

\begin{figure*}[t]
\begin{minipage}[b]{1\linewidth}
  \centering
  \centerline{\includegraphics[width=17.5cm]{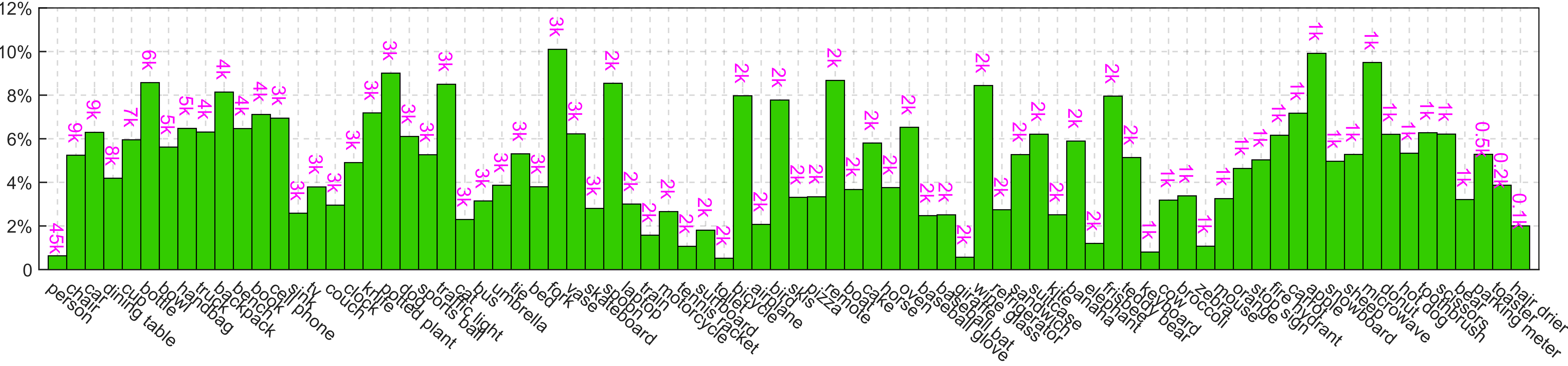}}
  \centerline{\small (a). The improvements over each class on MS-COCO.}
\end{minipage}
\begin{minipage}[b]{1\linewidth}
  \centering
  \centerline{\includegraphics[width=17.5cm]{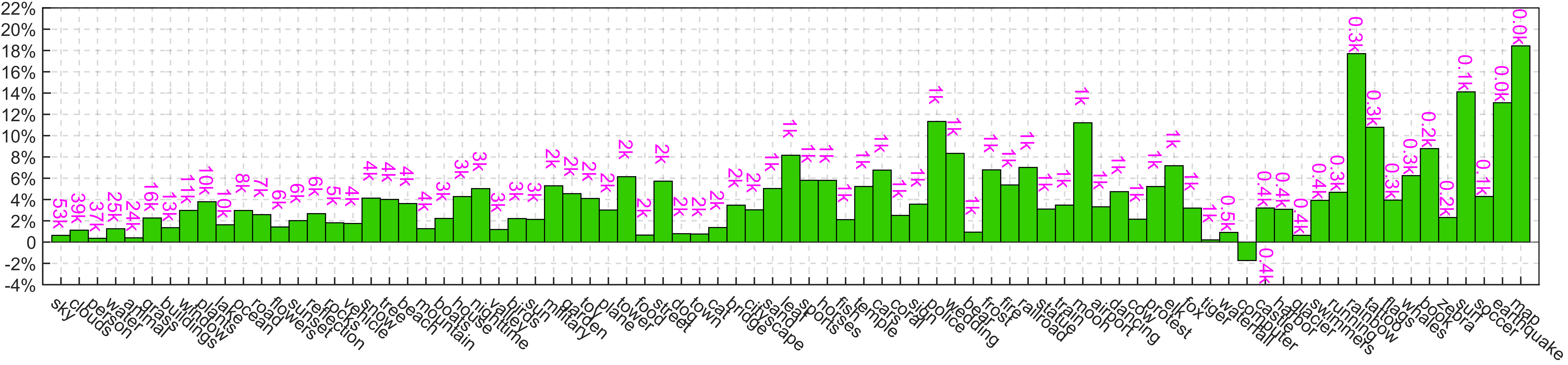}}
  \centerline{\small (b). The improvements over each concept on NUS-WIDE.}
\end{minipage}
\vspace{-15pt}
\caption{\small The improvements of S-Cls model over each class/concept on two datasets after knowledge distillation with our framework. ``\textcolor[rgb]{1.00,0.00,1.00}{*k}'' indicates the number (divided by 1000) of images including this class/concept. The classes/concepts in horizontal axis are sorted by the number ``\textcolor[rgb]{1.00,0.00,1.00}{*k}'' from large to small.}
\label{Fig:ap_class}
\vspace{-6pt}
\end{figure*}

The improvements over each class on two datasets are shown in Figure \ref{Fig:ap_class}. As it shows, on one hand, the improvements on MS-COCO are relatively even to the classes, while NUS-WIDE focuses on the classes in which the number of images is fewer. This demonstrates the effectiveness of our framework to mitigate the problem of data imbalance, since the NUS-WIDE dataset is very imbalanced (the number of images on ``sky'' is 53k while quite a few concepts only hold hundreds of images). On the other hand, the improvements on MS-COCO focus on small objects like ``bottle'', ``fork'', ``apple'' and so on, which may be difficult for the classification model to pay attention. This indicates the importance of semantic regions where T-WDet model is distilled into S-Cls model, which is shown in Figure \ref{Fig:results_show}. Moreover, on NUS-WIDE, the improvements focus on scenes (e.g., ``rainbow''), events (e.g., ``earthquake'') and objects (e.g., ``book''), which demonstrates the robustness of our framework to the types of labels.

\vspace{3pt}
\subsection{Ablation Study}
\noindent \textbf{Overall ablation study.}\ The results of overall ablation study on two datasets are summarised in Table \ref{tab:ablation_overall}. The T-WDet model achieves very good performance on MS-COCO while slightly better performance on NUS-WIDE. The main reason is that the clean object labels on MS-COCO are quite suitable for detection task while the noisy concept labels are not. Moreover, the S-Cls model is improved on both datasets after knowledge distillation by our framework, which verifies its effectiveness.

\noindent \textbf{Component-wise ablation study.} \ We also perform component-wise ablation study on MS-COCO to carefully evaluate the contribution of the critical components of our framework. The baseline is the VGG16-based S-Cls model trained with sigmoid logistic loss as Eq. \ref{Eq:loss_hard}. The results are summarised in Table \ref{tab:ablation_component}. It improves a little when directly applying the distillation methods proposed by \cite{KD1_nips15_goj}. After adding our class-aware distillation approach to baseline, the performance is improved much more (from 71.3 to 72.1). This demonstrates that our discriminative knowledge distillation at prediction level is superior than class-identical distillation \cite{KD1_nips15_goj}.

We then add feature-level knowledge distillation followed with class-aware distillation in the way of two-stage training. The performance is improved considerably (from 72.3 to 73.8) when we take NMS operation to all the proposals based on their objectness confidences, and it improves again when weighting the localized features with these confidences. This demonstrates that the objectness confidence obtained from T-WDet model is a reliable indicator of semantic of the proposal.

We also analyse our framework by the experiment using supervised detection results. Specifically, we replace the EB proposals input to T-WDet model by the detection results from Faster RCNN \cite{Faster-RCNN_nips15_skfj}. All the hyper-parameters are set to the same as the source code of \cite{Faster-RCNN_nips15_skfj}, which results in 100 proposals for each image. The results are summarised in Table \ref{tab:fasterRcnn}. As can be seen, the classification performance of T-WDet is improved from 78.6 to 81.1 when using the supervised detection results. After distillation with our framework, S-Cls model is improved to 76.3 compared with EB proposals to 74.6, where the gap is not obvious. This further demonstrates the effectiveness of our proposed framework.

\begin{table}[t]
  \centering
  \small
  \caption{\small Overall ablation study on two datasets (\%). ``w/'' and ``w/o'' indicate ``with'' and ``without'' knowledge distillation by the proposed framework, respectively.}
  \vspace{-7pt}
  \label{tab:ablation_overall}
  \begin{tabular}{l|ccc}
  \Xhline{1.2pt}
  \multirow{2}*{Dataset}&\multicolumn{3}{c}{mAP} \\
  \cline{2-4}
  &S-Cls (w/o)&T-WDet&S-Cls (w/) \\
  \Xhline{1.2pt}
  MS-COCO&70.9&78.6&74.6 \\
  NUS-WIDE&55.6&58.2&60.1 \\
  \Xhline{1.2pt}
  \end{tabular}
  \vspace{-5pt}
\end{table}
\begin{table}[t]
  \centering
  \small
  \caption{\small Component-wise ablation study (\%).}
  \vspace{-7pt}
  \label{tab:ablation_component}
  \begin{tabular}{l|c}
  \Xhline{1.2pt}
  Method&mAP \\
  \Xhline{1.2pt}
  Baseline (Sigmoid-Logistic)&70.9 \\
 \hline
  +Distillation \cite{KD1_nips15_goj}&71.3 \\
  +Class-aware distillation&72.1 \\
  +NMS proposals transfer+Class-aware transfer&73.8 \\
  +RoI-aware transfer+Class-aware transfer&74.6 \\
  \Xhline{1.2pt}
  \end{tabular}
  \vspace{-8pt}
\end{table}

\begin{table}[t]
  \centering
  \small
  \caption{\small The comparison of region proposals from EdgeBoxes \cite{WSD3_eccv14_cp} and Faster-RCNN \cite{Faster-RCNN_nips15_skfj} (\%).}
  \vspace{-5pt}
  \label{tab:fasterRcnn}
  \begin{tabular}{l|c}
  \Xhline{1.2pt}
  Method&mAP \\
  \Xhline{1.2pt}
  Baseline (Sigmoid-Logistic)&70.9 \\
  \hline
  T-WDet (EdgeBoxes \cite{WSD3_eccv14_cp})&78.6 \\
  S-Cls&74.6 \\
  \hline
  T-WDet (Faster RCNN \cite{Faster-RCNN_nips15_skfj})&81.1 \\
  S-Cls&76.3 \\
  \Xhline{1.2pt}
  \end{tabular}
  \vspace{-8pt}
\end{table}

\vspace{+5pt}
\section{Conclusion}

In this paper, a novel and efficient deep framework for multi-label image classification has been proposed. It boosts classification by distilling the unique knowledge from weakly-supervised detection (WSD) into classification with only image-level annotations. The proposed framework works with two steps. A WSD model is first developed, then an end-to-end knowledge distillation framework is constructed via feature-level distillation from RoIs and class-level distillation from predictions, where the WSD model is the \emph{teacher} model and the classification model is the \emph{student} model. The feature-level distillation from RoIs learns to perceiving semantic regions detected by the WSD model while class-level distillation aims at capturing class dependencies in the predictions of the WSD model. Thanks to this effective distillation, the classification model could be significantly improved without the WSD model in the test phase, thus resulting in high efficiency. Extensive experiments on two large-scale public datasets (MS-COCO and NUS-WIDE) show that the proposed framework outperforms the state-of-the-arts on both performance and efficiency. In the future, we will explore the complementary cues that could facilitate weakly-supervised detection to further boost the multi-label classification task.
\vspace{+17pt}
\\
\noindent {\normalsize \textbf{ACKNOWLEDGEMENT}}
\\
This work was supported by the National Natural Science Foundation of China under grant 91646207. 

\newpage
\bibliographystyle{ACM-Reference-Format}
\bibliography{sample-bibliography}

\end{document}